\documentclass[letterpaper]{article} 
\usepackage{aaai23}  
\usepackage{times}  
\usepackage{helvet}  
\usepackage{courier}  
\usepackage[hyphens]{url}  
\usepackage{graphicx} 
\usepackage{natbib}  
\usepackage{caption} 
\usepackage{algorithm}
\usepackage{algorithmic}

  \usepackage{amsmath}
  \usepackage{amsfonts} 
  \usepackage{bm}

  \usepackage{adjustbox}
  \usepackage{graphicx}
  \usepackage{dblfloatfix} 
  \usepackage{float}

  \usepackage{siunitx} 
  \usepackage{multirow}
  \usepackage{booktabs} 

\usepackage{newfloat}
\usepackage{listings}
\DeclareCaptionStyle{ruled}{labelfont=normalfont,labelsep=colon,strut=off} 
\floatstyle{ruled}
\newfloat{listing}{tb}{lst}{}
\floatname{listing}{Listing}
\title{Toward Face Biometric De-identification using Adversarial Examples}

\title{Toward Face Biometric De-identification using Adversarial Examples}
\author {
    Mahdi Ghafourian,\textsuperscript{\rm 1}
    Julian Fierrez,\textsuperscript{\rm 1}
    Luis F. Gomez,\textsuperscript{\rm 1}
    Ruben Vera-Rodriguez,\textsuperscript{\rm 1}
    Aythami Morales,\textsuperscript{\rm 1}
    Zohra Rezgui,\textsuperscript{\rm 2}
    Raymond Veldhuis,\textsuperscript{\rm 2}
}
\affiliations {
    \textsuperscript{\rm 1}Universidad Autonoma de Madrid\\
    \textsuperscript{\rm 2}University of Twente\\
    \textsuperscript{\rm 1}\{mahdi.ghafourian, julian.fierrez, luisf.gomez, ruben.vera, aythami.morales\}@uam.es
    
    \textsuperscript{\rm 2}\{z.rezgui, r.n.j.veldhuis\}@utwente.nl
}

\begin{document}

\maketitle

\begin{abstract}
The remarkable success of face recognition (FR) has endangered the privacy of internet users particularly in social media. Recently, researchers turned to use adversarial examples as a countermeasure. In this paper, we assess the effectiveness of using two widely known adversarial methods (BIM and ILLC) for de-identifying personal images. We discovered, unlike previous claims in the literature, that it is not easy to get a high protection success rate (suppressing identification rate) with imperceptible adversarial perturbation to the human visual system. Finally, we found out that the transferability of adversarial examples is highly affected by the training parameters of the network with which they are generated.
\end{abstract}


%

\section{Introduction}
Deep learning has evolved as a strong and efficient tool to be applied to a broad spectrum of complex learning problems that were difficult to solve using traditional machine learning \cite{krizhevsky2017imagenet, simonyan2014very}. The development of deep convolutional neural networks (CNNs) has been so revolutionary that today it can exceed human‐level performance. As a consequence, they are being extensively used in most of the recent day‐to‐day applications including face recognition. Now, face recognition (FR) systems have become an exceptionally accurate technology in identifying people from images \cite{schroff2015facenet, he2016deep}. While being useful, face recognition may invade the privacy of individuals when used to exploit and process illicitly their face images \cite{REF1, REF2} and videos \cite{REF3, REF4} found on the internet, particularly social media.

In recent years, several reports revealed unauthorized collections of large datasets of identified face data from social media. Reports on Cambridge Analytica \cite{CambridgeAnalytica} in 2018, and Clearview AI in 2020 \cite{hill2020secretive} are glaring examples of privacy leakage related to face biometrics. So far, the most common defense against this threat has been to set all social media profiles to ‘private’, allowing only chosen friends access to your images \cite{ledford2021assessment}. 

\begin{figure*}[!b]
 \centering 
 \includegraphics[trim={2cm 5cm 2cm 5cm},clip,width=180mm,scale=0.5]{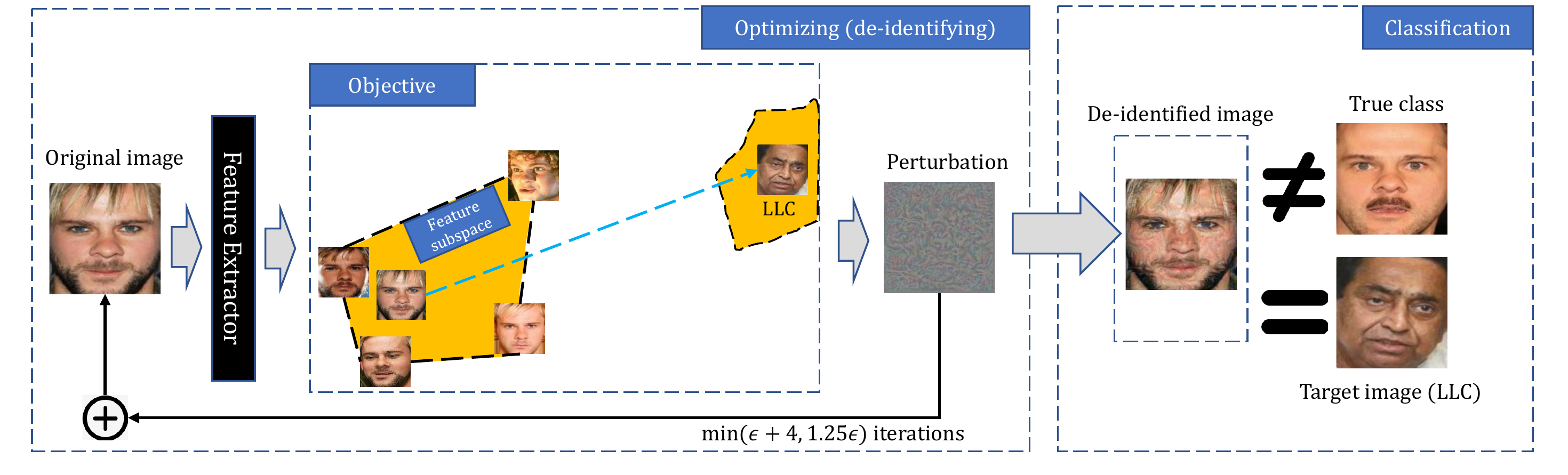}
 \caption{The overview of the targeted adversarial examples to de-idenify face images.}
 \label{fig:overview}
\end{figure*}
To mitigate these privacy threats, some studies \cite{shan2020fawkes, cherepanova2021lowkey, zhong2022opom, cilloni2022ulixes} turned to generate adversarial perturbations called cloaks to de-identify face biometrics in personal images before uploading them to social media. These perturbations are being generated by applying a very slight (imperceptible to human eyes) modification to the input and optimizing it to maximize the probability of misclassifcation by a machine learning classifier \cite{chakraborty2021survey, biggio2013evasion}. Using attacks to preserve privacy in biometrics has attracted attention \cite{ghafourian2022otb} which also includes adversarial examples. The goal of image cloaking for privacy protection is to suppress the identification rate of the subject while preserving the quality of their images \cite{REF5, REF6} keeping the adversarial perturbation imperceptible. 

In another line of work, instead of introducing imperceptible artifacts at the raw image level to harden automatic identification, one can operate at the feature level by disentangling there the identification information and reducing it while preserving other information of interest (e.g., facial emotions \cite{REF7}, soft biometrics \cite{REF8}, etc.) See the work by Morales et al. \cite{REF9} and the references therein for further information in this line.

In the present paper, we conduct an experimental evaluation of the effectiveness of two popular adversarial methods, i.e. Basic Iterative Method (BIM) and Iterative Least Likely Class (ILLC) \cite{kurakin2018adversarial}, for de-identifying face biometrics in personal photos at the raw image level. In particular, we focussed on the transferability of the de-identified face biometrics across different classifiers. To this end, we used three popular pre-trained face recognition models including (\textit{FaceNet}, \textit{ResNet-50}, and \textit{SENet-50}) interchangeably to create an adversarial example by one model and defend against it using all three models. To fully demonstrate the performance of the experimented adversarial methods for privacy preservation, we report the protection success rate of the generated examples on the defender networks at various noise budgets and classification rates.

By analyzing the quantitative results of BIM and ILLC methods, we obtained some important findings. First, it is not likely to obtain a high protection success rate together with quite imperceptible adversarial perturbation. In particular, when it comes to black-box scenarios and any preprocessing (e.g. image compression, resizing) that affects the adversarial trigger, this goal would be ambitious. Second, we discuss that the definition of feature embeddings of the adversarial class are highly dependent on the other training classes in the attacker network. Therefore, the transferability of generated adversarial examples (i.e. de-identified personal images) conforms with the similarity of the attacker network to that of the defender in terms of training parameters. Third, unlike our expectation, although the BIM method is an untargeted method (i.e. adversarial method without an specific target), it is more protective than the targeted ILLC method.

\section{Protection model}
In this section, we introduce the protector's goal, capabilities, and knowledge under which the de-identified samples are generated. Since the goal of our study is to preserve the privacy using adversarial examples, we call the party who generates the examples the protector and the party whose network is used for classifying the examples, the invader. For a better understanding of the paper, we provide definitions from their original sources with which we conducted our experiments. Therefore, in the remaining of the paper, we use the following notations:
\begin{itemize}
  \item $x$:  the input face biometric of the identity who wants to be de-identified. It is an RGB image in the shape of a 3D tensor $(width \times height \times depth) $ whose values range is in [0, 255].
  \item $x_{adv}$:  the adversarial example (i.e. de-identified image) for $x$.
  \item $y_{true}$: the true class label for the image $x$.
  \item $y_{target}$: the target class label that the defender is trying to optimize the input image to fool the attacker classifier with, in our case the least likely class ($y_{LLC}$).
  \item $\epsilon$ the noise budget to add to one pixel of $x$.
  \item $C(x)$: it denotes the classifier $ C(x): X \rightarrow Y $ where $x \in X \subset \mathbb{R}^{d}$, and $y = \{ 1,2, \cdots , N \}$ with $N$ being the total number of classes.
  \item $J(x, y_{target})$: the cross-entropy cost function for computing the loss of $x$ given the target class label $y_{target}$.
  \item Clip$_{\epsilon } \{ x_{adv} \} $: clipping function to confine the alteration of each pixel in the de-identified image $x_{adv}$ to the noise budget $\epsilon$ to keep the result in the $L_{P}$ $ \epsilon$-neighbourhood of the input image $x$.
\end{itemize}

\subsection{Protector's goal}
The goal of the protector is to craft an adversarial perturbation to hinder automatic face recognition (face de-identification) while keeping the visual appearance. To this end, the protector adds a small perturbation measured by $L_{P}$ norm to the original face biometric in a specific number of iterations. For the adversarial method we used, the upper bound of this number of iterations is determined by $\min( \epsilon + 4, 1.25 \epsilon)$. In general adversarial methods are divided into two categories: 
\begin{itemize}
  \item \textbf{Untargeted} the aim of adversarial examples crafted by these methods is to send away the classification result from the true class $y_{true}$ to mislead the classifier as $C(x_{adv}) \neq y_{true}$.
  \item \textbf{Targeted} the goal of adversarial examples crafted by these methods is to misdirect the classification result to the desired target class $y_{target}$ as $C(x_{adv}) = y_{target}$ (see Figure \ref{fig:overview}).
\end{itemize}

\subsection{Protector's capability}
To achieve the goal, the protector crafts de-identified face biometrics with constrained perturbation. To this end, the adversarial examples generated by this approach need to satisfy $\lVert x_{adv} - x \rVert _{p} \leq \epsilon$ to mislead the model of the privacy invader. Therefore, the protector is able to conduct the following optimization problems in the aforementioned number of iterations according to the method he adopts. Regarding the untargeted methods, the protector generates the de-identified face by maximizing the cost function $J(x_{adv}, y_{true})$ as:

\begin{equation} \label{eq1}
\begin{split}
x_{adv} = \operatorname*{argmax}_{x_{adv}: {\lVert x_{adv} - x \rVert _{p} \leq \epsilon}} J(x_{adv}, y_{true})
\end{split}
\end{equation}
while for the targeted method, de-identified face images are crafted by minimizing the cost function $J(x_{adv}, y_{target})$ as:

\begin{equation} \label{eq2}
\begin{split}
x_{adv} = \operatorname*{argmin}_{x_{adv}: {\lVert x_{adv} - x \rVert _{p} \leq \epsilon}} J(x_{adv}, y_{target})
\end{split}
\end{equation}

\subsection{Protector's knowledge}
Similar to the real-world scenarios, we conducted our assessment in a black-box setting. In black-box attacks, it is assumed that the protector has no prior knowledge of the invader's network or its parameters. With this assumption, the protector can only acquire the classification output of the invader model. Therefore, in an oracle attack, the protector evaluates the protection success rate by providing crafted inputs with various perturbation budgets. However, the protector can use the same dataset for generating adversarial examples with which the invader's model has been trained.

\section{Generating de-identified faces}
\begin{figure*}[!t]
 \centering 
 \includegraphics[trim={9.5cm 1cm 0 1cm},clip,width=310mm,scale=0.5]{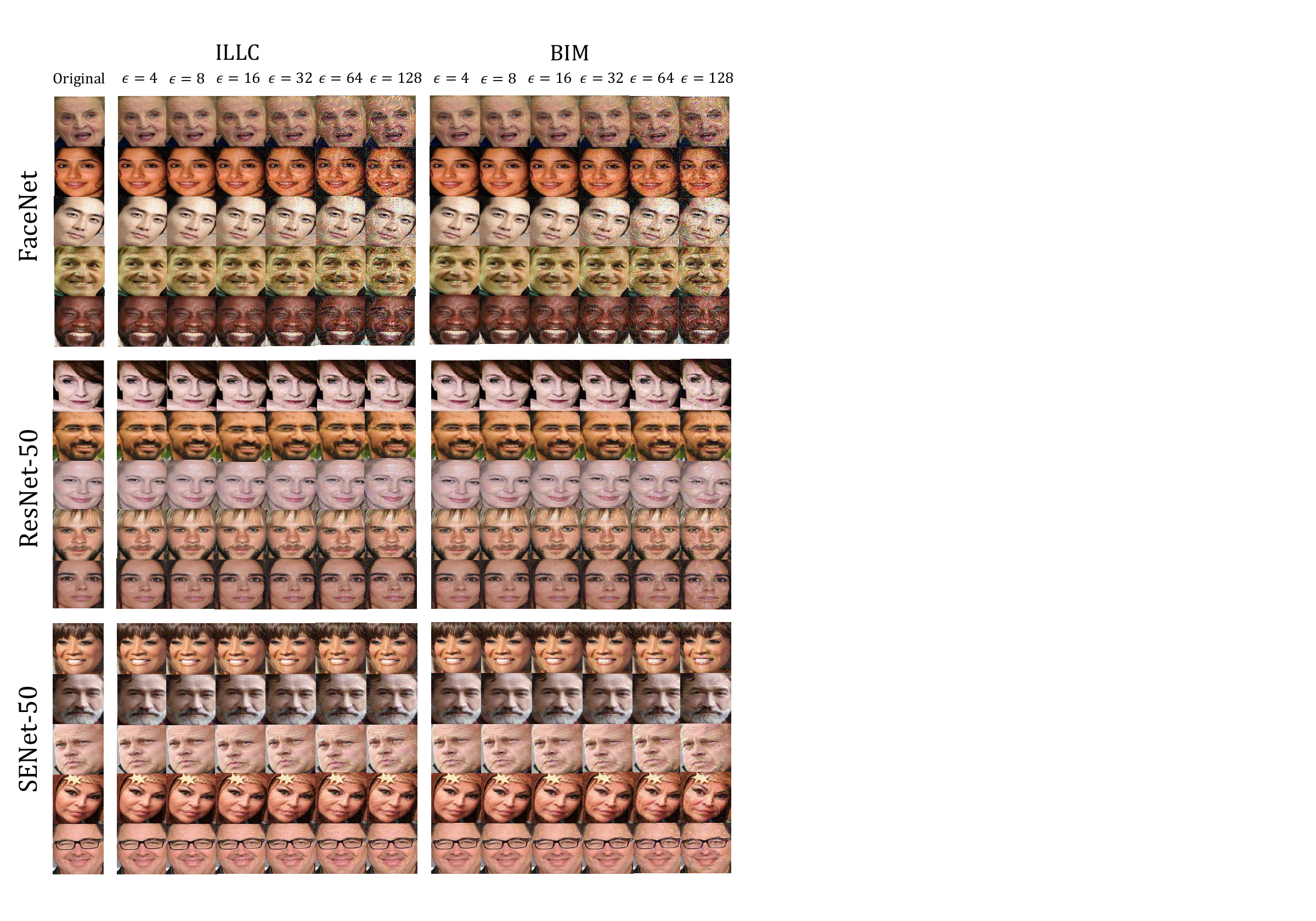}
 \caption{Example of de-identified face images crafted by ILLC and BIM methods for all the models with various perturbation budgets ($\epsilon$). }
 \label{fig:de-identified}
\end{figure*}

The aim of de-identification on face biometrics is to preserve the privacy of subjects by protecting their true identity against unwanted face identifications. To this end, the use of adversarial perturbations through a technique called Image Cloaking has been proposed recently. In this line of work, Shan et al. \cite{shan2020fawkes} proposed a method called Fawkes, harnessing image cloaking technique to reduce the effectiveness of face recognition software while preserving the quality of the image to human eyes. 
In this method, the target face recognition model needs to be trained with the clocked images. The author of Fawkes reports $95\%$ protection success rate for top-1 identification applied to commercial off-the-shelf face recognition. Another similar work called LowKey \cite{cherepanova2021lowkey} did the image cloaking by updating $x_{adv}$ iteratively adding the sign of gradient toward the maximization objective. They applied Gaussian smoothing to maintain the quality of the image and could reduce the accuracy of Amazon Rekognition \cite{AmazonRekognition} to $32.5\%$ (i.e. $67.5\%$ protection rate). In this paper, we generate de-identified face images with various perturbation budgets using BIM and ILLC adversarial methods as it is shown in Figure \ref{fig:de-identified}.

\subsection{Basic Iterative Method (BIM)}
According to \cite{goodfellow2014explaining}, the easiest way to generate an adversarial image is to find the perturbation that maximizes the cost function with respect to $L_{ \infty }$ constraint with just one back-propagation iteration (FGSM method). Later, \cite{kurakin2018adversarial} extended this method by doing back-propagation iteratively while it is clipping values changes in pixels after each iteration to keep the alteration to the $ \epsilon$-neighbourhood of the original image. This method is called BIM and the adversarial image in each iteration is crafted as below:
\begin{equation} \label{eq3}
\begin{split}
x_{adv}^{(i+1)} = \textrm{Clip}_{\epsilon } \{ x_{adv}^{(i)} + \alpha . \textrm{sign}(\nabla_{x_{adv}^{(i)}} J(x_{adv}^{(i)}, y_{true}) \}
\end{split}
\end{equation}
where $\alpha$ is the step size and $x_{adv}^{(0)} = x$ at the initialization of BIM method. By maximizing the cost $J$ in this iterative way, the classification result of the de-identified face image $x_{adv}$ would lie far from the original image $x$.

\subsection{Iterative Least Likely Class (ILLC)}
Unlike BIM, the only difference of this method is to reduce the cost but toward a specific target. In this case, the target is the least likely class when the original image is classified. As a result, the crafted de-identified face will be closer to another person in the classification database. The effectiveness of this method for de-identification relies on the dissimilarity rate of all the subjects in the training dataset. This method is also an iterative method initiated with $x_{adv}^{(0)} = x$ and the adversarial image in each iteration is crafted as:
\begin{equation} \label{eq4}
\begin{split}
x_{adv}^{(i+1)} = \textrm{Clip}_{\epsilon } \{ x_{adv}^{(i)} - \alpha . \textrm{sign}(\nabla_{x_{adv}^{(i)}} J(x_{adv}^{(i)}, y_{LLC}) \}
\end{split}
\end{equation}

\section{Evaluation}
\begin{figure*}[!b]
 \centering 
 \includegraphics[trim={0cm 2cm 0cm 2cm},clip,width=170mm,scale=0.5]{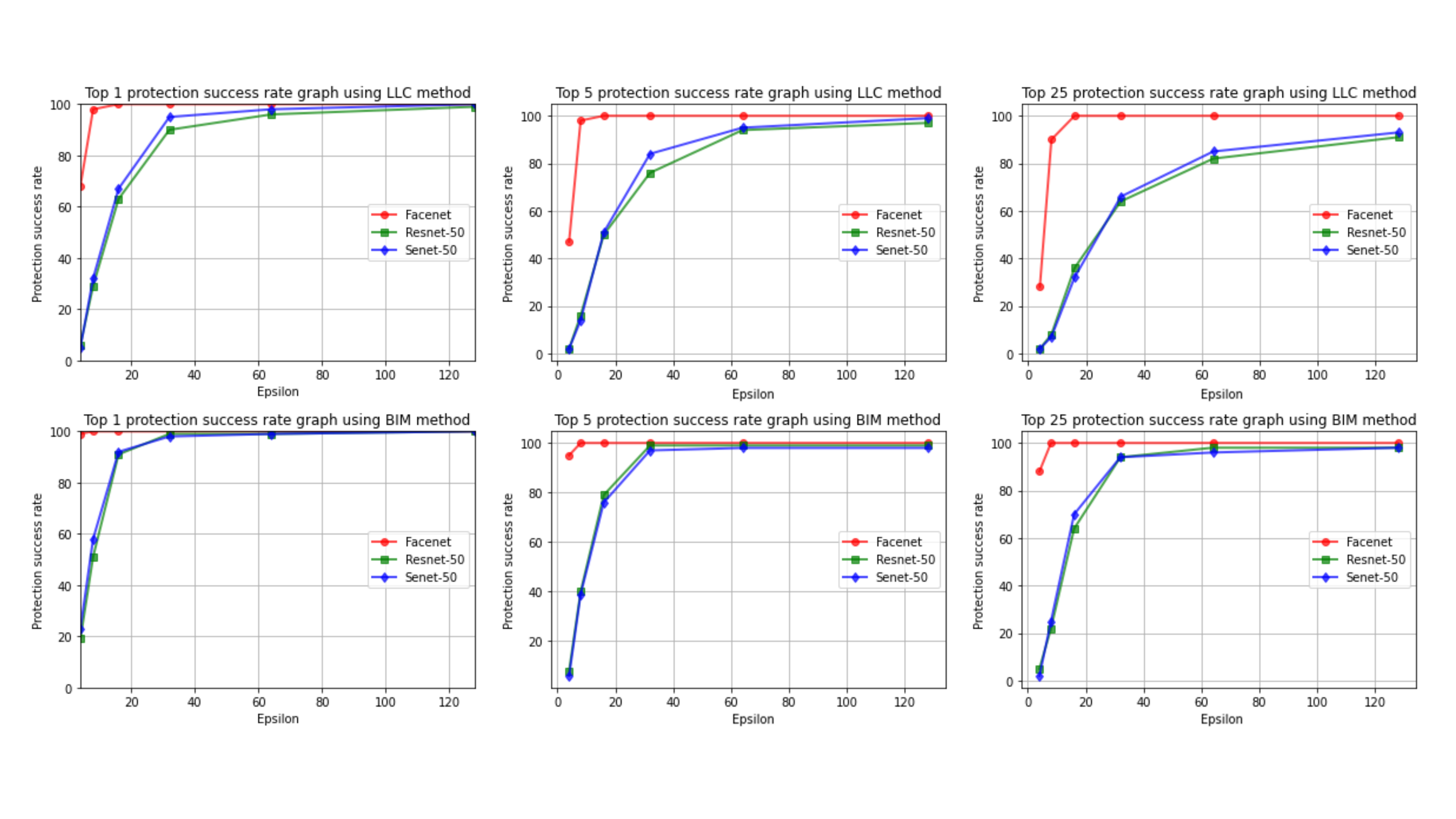}
 \caption{Protection Success Rate (PSR) as the perturbation budget ($\epsilon$) increases for adversarial examples crafted using FaceNet. First row: using ILLC method with different accuracy (left to right: Top1, Top5, Top25). Second row: idem using BIM method.}
 \label{fig:PSR_Facenet}
\end{figure*}

\begin{figure*}[!b]
 \centering 
 \includegraphics[trim={0cm 2cm 0cm 2cm},clip,width=170mm,scale=0.5]{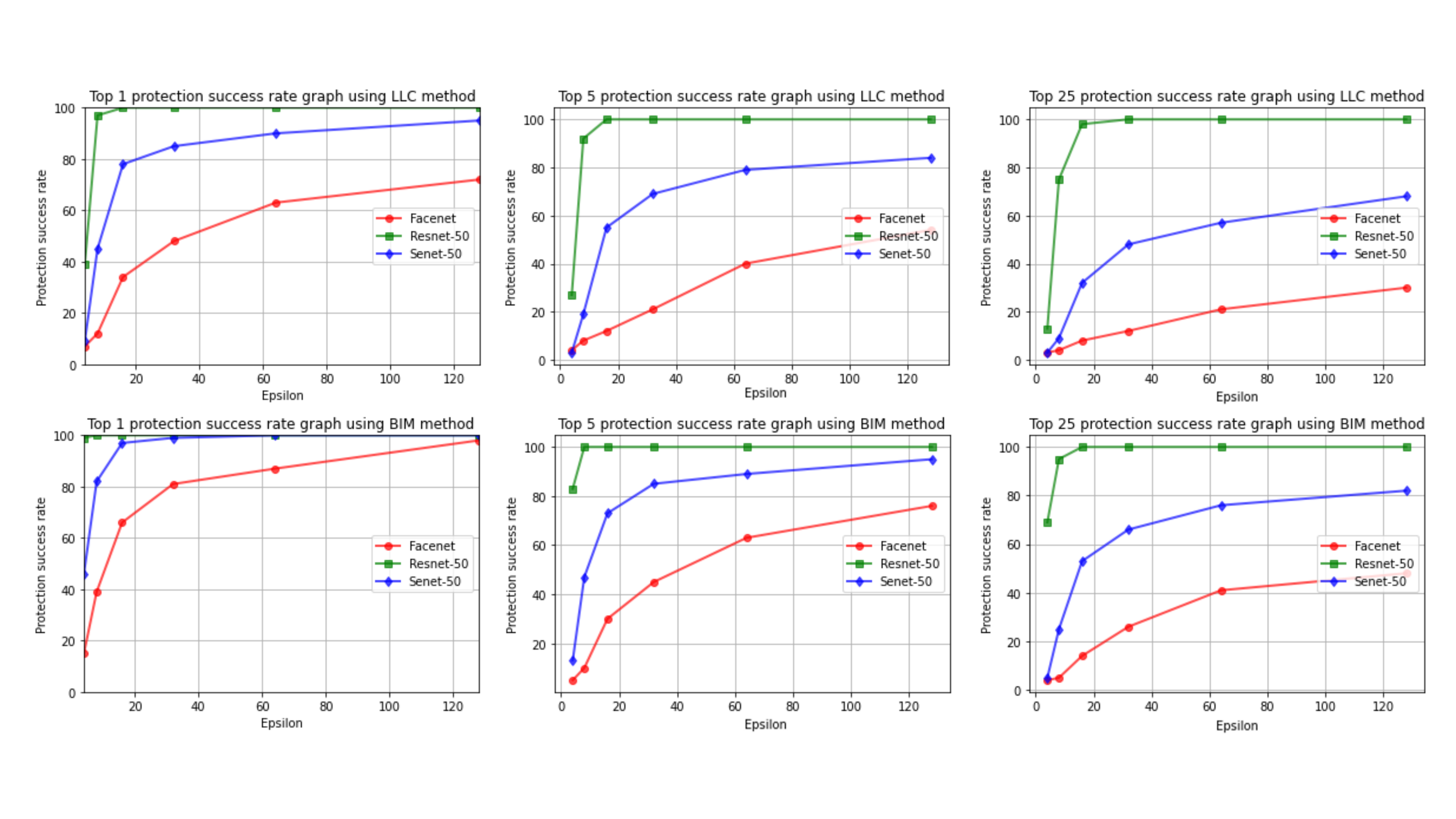}
 \caption{Protection Success Rate (PSR) as the perturbation budget ($\epsilon$) increases for adversarial examples crafted using ResNet. First row: using ILLC method with different accuracy (left to right: Top1, Top5, Top25). Second row: idem using BIM method.}
 \label{fig:PSR_Resnet}
\end{figure*}

\subsection{Evaluation metric}
So far, the most common metric that has been used to evaluate the performance of adversarial examples is transferability. This metric denotes that the examples produced to deceive a particular model can be used to deceive other models regardless of the underlying architecture. To estimate the transferability of the generated adversarial examples we use the Protection Success Rate (PSR) also called the suppressing identification rate. In our case, PSR is the misclassification rate of the de-identified faces by the target classifier. Thus, given the adversarial method $\textrm{Adv}_{\epsilon}$ to generate the de-identified face image as $x_{adv} = \textrm{Adv}_{\epsilon}(x)$ for the input face $x$ under the constraints of perturbation budget $\epsilon$ and $l_{p}$-norm, and target classifier $C(x)$, the PSR is defined as: 
\begin{equation} \label{eq5}
\begin{split}
\textrm{PSR}(\textrm{Adv}_{\epsilon}, C) = 100 - (\frac{100}{N}\Sigma_{i=1}^{N}1(C(\textrm{Adv}_{\epsilon}(x_i)) = y_{true})) 
\end{split}
\end{equation}
where $N$ is the number of test samples and $1(.)$ is the indicator function. The higher the PSR, the more resilient the example is to be identified by the target classifier.

\begin{figure*}[!t]
 \centering 
 \includegraphics[trim={0cm 1.5cm 0cm 2cm},clip,width=170mm,scale=0.5]{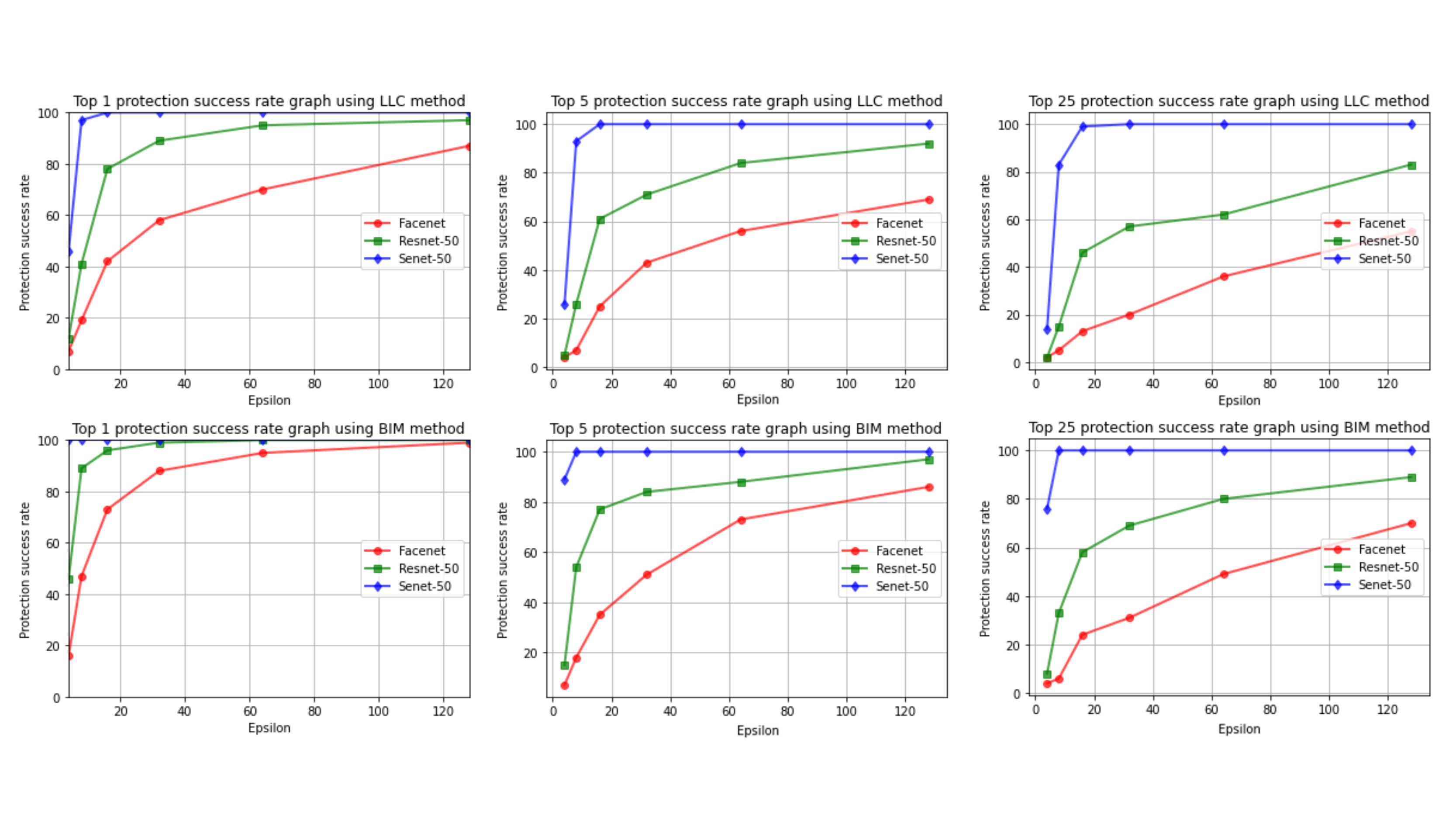}
 \caption{Protection Success Rate (PSR) as the perturbation budget ($\epsilon$) increases for adversarial examples crafted using SENet. First row: using ILLC method with different accuracy (left to right: Top1, Top5, Top25). Second row: idem using BIM method.}
 \label{fig:PSR_Senet}
\end{figure*}

\subsection{Evaluation settings}
Our experiments are divided into two phases. \textit{Generating the de-identified image} of the input face in the source network by the protector, and \textit{classifying the example} in target networks to evaluate the Protection Success Rate (PSR). To this end, we used three widely used pre-trained face recognition models (all trained on the VGGFace2 dataset \cite{21cao2018vggface2}): FaceNet \cite{schroff2015facenet}, ResNet-50 \cite{he2016deep}, SENet-50 \cite{hu2018squeeze}. 
We start the process of generating de-identified faces in the source network as follows:
First, $N$ random subjects from the VGGFace2 dataset are selected. These are subjects that we are going to protect their identity. Second, the perturbation budget $\epsilon$ is picked from the  $set_{\epsilon} = \{ 4, 8, 16, 32, 64, 128 \}$ \cite{kurakin2018adversarial}. In terms of the transferability, what is important in our experiments is to assess the proportion of Protection Success Rate to the image quality degradation. The ideal output is to achieve the largest PSR using the smallest possible perturbation budget. Third, the number of iterations for optimizing the input face toward the adversarial goal is calculated as $n_{iter} = \min(\epsilon + 4, 1.25 \times \epsilon)$. Finally, for every \textit{Model} $\in$ \{\textit{FaceNet, ResNet-50, SENet-50}\} as source network, for each random input face $x \in \{ x_i \}_{i=1}^N$, and for every $\epsilon \in set_{\epsilon}$, we iterate the input image $x$ by $n_{iter}$ doing backpropagation toward $y_{target}$ for $ILLC$ method and $y_{true}$ for BIM method. Some examples of de-identified face images regarding both adversarial methods for each $\epsilon \in set_{\epsilon}$ are depicted in Figure \ref{fig:de-identified}.
Once the de-identified face is crafted, for each \textit{Model} $\in$ \{\textit{FaceNet, ResNet-50, SENet-50}\} as target networks, we assess the protection success rate of the crafted examples via the following steps:
First, the face is extracted using MTCNN \cite{zhang2016joint} to make sure that the perturbation hasn't made the face undetectable. Second, the detected face is fed to the classifier of the selected \textit{Model}. Third, based on the classification maximum probability, we compute Top1, Top5, Top10, Top25, Top50 where $C(\textrm{Adv}_{\epsilon}(x_i)) = y_{true}$. Finally, for each top, we calculate PSR according to Equation \ref{eq5}. The resulting PSR for the $n_{iter}$ corresponding to each $\epsilon \in set_{\epsilon}$ for FaceNet, ResNet-50, and SENet-50 is depicted in Figures \ref{fig:PSR_Facenet}, \ref{fig:PSR_Resnet}, \ref{fig:PSR_Senet}  respectively. 

\begin{table*}[t]
\caption{Comparison of BIM and ILLC methods in terms of protection success rate using different models with various noise budgets to generate de-identified faces. We report Top-1, 5, 10, 25, and 50 protection success rate under 1:N identification setting. The values are in percentage and the higher
protection success rate is better.}
\label{tab:ApprochesComparison}
\resizebox{17cm}{!}{%
\begin{tabular}{@{}l|c|l|ccccccccccccccc@{}}
\multicolumn{1}{c|}{\multirow{3}{*}{\textbf{Source}}} & \multirow{3}{*}{\textbf{Method}} & \multirow{3}{*}{\textbf{Noise Budget}} & \multicolumn{15}{c}{\textbf{Target}}                                                                                        \\ \cmidrule(l){4-18} 
\multicolumn{1}{c|}{}                                 &                                  &                                        & \multicolumn{5}{c|}{\textbf{FaceNet}}                                                                                                                                                      & \multicolumn{5}{c|}{\textbf{ResNet-50}}                                                                                                                                                    & \multicolumn{5}{c}{\textbf{SENet-50}}                                                                                                                                                     \\ \cmidrule(l){4-18} 
\multicolumn{1}{c|}{}                                 &                                  &                                        & \multicolumn{1}{l}{\textbf{Top-1}} & \multicolumn{1}{l}{\textbf{Top-5}} & \multicolumn{1}{l}{\textbf{Top-10}} & \multicolumn{1}{l}{\textbf{Top-25}} & \multicolumn{1}{l|}{\textbf{Top-50}} & \multicolumn{1}{l}{\textbf{Top-1}} & \multicolumn{1}{l}{\textbf{Top-5}} & \multicolumn{1}{l}{\textbf{Top-10}} & \multicolumn{1}{l}{\textbf{Top-25}} & \multicolumn{1}{l|}{\textbf{Top-50}} & \multicolumn{1}{l}{\textbf{Top-1}} & \multicolumn{1}{l}{\textbf{Top-5}} & \multicolumn{1}{l}{\textbf{Top-10}} & \multicolumn{1}{l}{\textbf{Top-25}} & \multicolumn{1}{l}{\textbf{Top-50}} \\ \midrule
\multirow{12}{*}{\textbf{FaceNet}}                    & \multirow{6}{*}{\textbf{ILLC}}   & $\epsilon$ = 4                                  & 68.0                               & 47.0                               & 36.0                                & 28.0                                & \multicolumn{1}{c|}{22.0}            & 6.0                                & 2.0                                & 2.0                                 & 2.0                                 & \multicolumn{1}{c|}{2.0}             & 5.0                                & 2.0                                & 2.0                                 & 2.0                                 & 2.0                                 \\
                                                      &                                  & $\epsilon$ = 8                                  & 98.0                               & 98.0                               & 92.0                                & 90.0                                & \multicolumn{1}{c|}{89.0}            & 29.0                               & 16.0                               & 12.0                                & 8.0                                 & \multicolumn{1}{c|}{7.0}             & 32.0                               & 14.0                               & 8.0                                 & 7.0                                 & 4.0                                 \\
                                                      &                                  & $\epsilon$ = 16                                 & 100.0                              & 100.0                              & 100.0                               & 100.0                               & \multicolumn{1}{c|}{100.0}           & 63.0                               & 50.0                               & 43.0                                & 36.0                                & \multicolumn{1}{c|}{28.0}            & 67.0                               & 51.0                               & 47.0                                & 32.0                                & 26.0                                \\
                                                      &                                  & $\epsilon$ = 32                                 & 100.0                              & 100.0                              & 100.0                               & 100.0                               & \multicolumn{1}{c|}{100.0}           & 90.0                               & 76.0                               & 71.0                                & 64.0                                & \multicolumn{1}{c|}{60.0}            & 95.0                               & 84.0                               & 80.0                                & 66.0                                & 56.0                                \\
                                                      &                                  & $\epsilon$ = 64                                 & 100.0                              & 100.0                              & 100.0                               & 100.0                               & \multicolumn{1}{c|}{100.0}           & 96.0                               & 94.0                               & 91.0                                & 82.0                                & \multicolumn{1}{c|}{78.0}            & 98.0                               & 95.0                               & 92.0                                & 85.0                                & 78.0                                \\
                                                      &                                  & $\epsilon$ = 128                                & 100.0                              & 100.0                              & 100.0                               & 100.0                               & \multicolumn{1}{c|}{100.0}           & 99.0                               & 97.0                               & 95.0                                & 91.0                                & \multicolumn{1}{c|}{88.0}            & 100.0                              & 99.0                               & 99.0                                & 93.0                                & 89.0                                \\ \cmidrule(lr){2-2}
                                                      & \multirow{6}{*}{\textbf{BIM}}    & $\epsilon$ = 4                                  & 99.0                               & 95.0                               & 93.0                                & 88.0                                & \multicolumn{1}{c|}{80.0}            & 19.0                               & 8.0                                & 6.0                                 & 5.0                                 & \multicolumn{1}{c|}{3.0}             & 23.0                               & 6.0                                & 3.0                                 & 2.0                                 & 2.0                                 \\
                                                      &                                  & $\epsilon$ = 8                                  & 100.0                              & 100.0                              & 100.0                               & 100.0                               & \multicolumn{1}{c|}{100.0}           & 51.0                               & 40.0                               & 33.0                                & 22.0                                & \multicolumn{1}{c|}{20.0}            & 58.0                               & 39.0                               & 33.0                                & 25.0                                & 20.0                                \\
                                                      &                                  & $\epsilon$ = 16                                 & 100.0                              & 100.0                              & 100.0                               & 100.0                               & \multicolumn{1}{c|}{100.0}           & 91.0                               & 79.0                               & 71.0                                & 64.0                                & \multicolumn{1}{c|}{59.0}            & 92.0                               & 76.0                               & 75.0                                & 70.0                                & 61.0                                \\
                                                      &                                  & $\epsilon$ = 32                                 & 100.0                              & 100.0                              & 100.0                               & 100.0                               & \multicolumn{1}{c|}{100.0}           & 99.0                               & 99.0                               & 98.0                                & 94.0                                & \multicolumn{1}{c|}{89.0}            & 98.0                               & 97.0                               & 96.0                                & 94.0                                & 91.0                                \\
                                                      &                                  & $\epsilon$ = 64                                 & 100.0                              & 100.0                              & 100.0                               & 100.0                               & \multicolumn{1}{c|}{100.0}           & 99.0                               & 99.0                               & 98.0                                & 98.0                                & \multicolumn{1}{c|}{95.0}            & 99.0                               & 98.0                               & 97.0                                & 96.0                                & 96.0                                \\
                                                      &                                  & $\epsilon$ = 128                                & 100.0                              & 100.0                              & 100.0                               & 100.0                               & \multicolumn{1}{c|}{100.0}           & 100.0                              & 98.0                               & 98.0                                & 98.0                                & \multicolumn{1}{c|}{97.0}            & 100.0                              & 98.0                               & 98.0                                & 98.0                                & 98.0                                \\ \midrule
\multirow{12}{*}{\textbf{ResNet-50}}                  & \multirow{6}{*}{\textbf{ILLC}}   & $\epsilon$ = 4                                  & 7.0                                & 4.0                                & 4.0                                 & 3.0                                 & \multicolumn{1}{c|}{3.0}             & 39.0                               & 27.0                               & 19.0                                & 13.0                                & \multicolumn{1}{c|}{11.0}            & 9.0                                & 3.0                                & 3.0                                 & 3.0                                 & 2.0                                 \\
                                                      &                                  & $\epsilon$ = 8                                  & 12.0                               & 8.0                                & 5.0                                 & 4.0                                 & \multicolumn{1}{c|}{3.0}             & 97.0                               & 92.0                               & 84.0                                & 75.0                                & \multicolumn{1}{c|}{67.0}            & 45.0                               & 19.0                               & 13.0                                & 9.0                                 & 8.0                                 \\
                                                      &                                  & $\epsilon$ = 16                                 & 34.0                               & 12.0                               & 10.0                                & 8.0                                 & \multicolumn{1}{c|}{6.0}             & 100.0                              & 100.0                              & 100.0                               & 98.0                                & \multicolumn{1}{c|}{97.0}            & 78.0                               & 55.0                               & 42.0                                & 32.0                                & 24.0                                \\
                                                      &                                  & $\epsilon$ = 32                                 & 48.0                               & 21.0                               & 17.0                                & 12.0                                & \multicolumn{1}{c|}{9.0}             & 100.0                              & 100.0                              & 100.0                               & 100.0                               & \multicolumn{1}{c|}{98.0}            & 85.0                               & 69.0                               & 59.0                                & 48.0                                & 36.0                                \\
                                                      &                                  & $\epsilon$ = 64                                 & 63.0                               & 40.0                               & 27.0                                & 21.0                                & \multicolumn{1}{c|}{15.0}            & 100.0                              & 100.0                              & 100.0                               & 100.0                               & \multicolumn{1}{c|}{100.0}           & 90.0                               & 79.0                               & 70.0                                & 57.0                                & 47.0                                \\
                                                      &                                  & $\epsilon$ = 128                                & 72.0                               & 54.0                               & 44.0                                & 30.0                                & \multicolumn{1}{c|}{25.0}            & 100.0                              & 100.0                              & 100.0                               & 100.0                               & \multicolumn{1}{c|}{100.0}           & 95.0                               & 84.0                               & 78.0                                & 68.0                                & 60.0                                \\ \cmidrule(lr){2-2}
                                                      & \multirow{6}{*}{\textbf{BIM}}    & $\epsilon$ = 4                                  & 15.0                               & 5.0                                & 5.0                                 & 4.0                                 & \multicolumn{1}{c|}{4.0}             & 99.0                               & 83.0                               & 74.0                                & 69.0                                & \multicolumn{1}{c|}{57.0}            & 46.0                               & 13.0                               & 9.0                                 & 5.0                                 & 4.0                                 \\
                                                      &                                  & $\epsilon$ = 8                                  & 39.0                               & 10.0                               & 7.0                                 & 5.0                                 & \multicolumn{1}{c|}{4.0}             & 100.0                              & 100.0                              & 99.0                                & 95.0                                & \multicolumn{1}{c|}{91.0}            & 82.0                               & 47.0                               & 38.0                                & 25.0                                & 18.0                                \\
                                                      &                                  & $\epsilon$ = 16                                 & 66.0                               & 30.0                               & 23.0                                & 14.0                                & \multicolumn{1}{c|}{8.0}             & 100.0                              & 100.0                              & 100.0                               & 100.0                               & \multicolumn{1}{c|}{99.0}            & 97.0                               & 73.0                               & 65.0                                & 53.0                                & 45.0                                \\
                                                      &                                  & $\epsilon$ = 32                                 & 81.0                               & 45.0                               & 36.0                                & 26.0                                & \multicolumn{1}{c|}{21.0}            & 100.0                              & 100.0                              & 100.0                               & 100.0                               & \multicolumn{1}{c|}{99.0}            & 99.0                               & 85.0                               & 75.0                                & 66.0                                & 55.0                                \\
                                                      &                                  & $\epsilon$ = 64                                 & 87.0                               & 63.0                               & 49.0                                & 41.0                                & \multicolumn{1}{c|}{30.0}            & 100.0                              & 100.0                              & 100.0                               & 100.0                               & \multicolumn{1}{c|}{100.0}           & 100.0                              & 89.0                               & 86.0                                & 76.0                                & 59.0                                \\
                                                      &                                  & $\epsilon$ = 128                                & 98.0                               & 76.0                               & 62.0                                & 48.0                                & \multicolumn{1}{c|}{41.0}            & 100.0                              & 100.0                              & 100.0                               & 100.0                               & \multicolumn{1}{c|}{100.0}           & 100.0                              & 95.0                               & 90.0                                & 82.0                                & 70.0                                \\ \midrule
\multirow{12}{*}{\textbf{SENet-50}}                   & \multirow{6}{*}{\textbf{ILLC}}   & $\epsilon$ = 4                                  & 7.0                                & 4.0                                & 4.0                                 & 2.0                                 & \multicolumn{1}{c|}{2.0}             & 12.0                               & 5.0                                & 4.0                                 & 2.0                                 & \multicolumn{1}{c|}{2.0}             & 46.0                               & 26.0                               & 19.0                                & 14.0                                & 12.0                                \\
                                                      &                                  & $\epsilon$ = 8                                  & 19.0                               & 7.0                                & 5.0                                 & 5.0                                 & \multicolumn{1}{c|}{4.0}             & 41.0                               & 26.0                               & 21.0                                & 15.0                                & \multicolumn{1}{c|}{13.0}            & 97.0                               & 93.0                               & 91.0                                & 83.0                                & 80.0                                \\
                                                      &                                  & $\epsilon$ = 16                                 & 42.0                               & 25.0                               & 21.0                                & 13.0                                & \multicolumn{1}{c|}{9.0}             & 78.0                               & 61.0                               & 55.0                                & 46.0                                & \multicolumn{1}{c|}{35.0}            & 100.0                              & 100.0                              & 100.0                               & 100.0                               & 98.0                                \\
                                                      &                                  & $\epsilon$ = 32                                 & 58.0                               & 43.0                               & 36.0                                & 2.0                                 & \multicolumn{1}{c|}{18.0}            & 89.0                               & 71.0                               & 68.0                                & 57.0                                & \multicolumn{1}{c|}{46.0}            & 100.0                              & 100.0                              & 100.0                               & 100.0                               & 99.0                                \\
                                                      &                                  & $\epsilon$ = 64                                 & 70.0                               & 56.0                               & 47.0                                & 36.0                                & \multicolumn{1}{c|}{29.0}            & 95.0                               & 84.0                               & 72.0                                & 62.0                                & \multicolumn{1}{c|}{61.0}            & 100.0                              & 100.0                              & 100.0                               & 100.0                               & 100.0                               \\
                                                      &                                  & $\epsilon$ = 128                                & 87.0                               & 69.0                               & 62.0                                & 55.0                                & \multicolumn{1}{c|}{42.0}            & 97.0                               & 92.0                               & 87.0                                & 83.0                                & \multicolumn{1}{c|}{74.0}            & 100.0                              & 100.0                              & 100.0                               & 100.0                               & 100.0                               \\ \cmidrule(lr){2-2}
                                                      & \multirow{6}{*}{\textbf{BIM}}    & $\epsilon$ = 4                                  & 16.0                               & 7.0                                & 5.0                                 & 4.0                                 & \multicolumn{1}{c|}{4.0}             & 46.0                               & 15.0                               & 10.0                                & 8.0                                 & \multicolumn{1}{c|}{6.0}             & 100.0                              & 89.0                               & 83.0                                & 76.0                                & 71.0                                \\
                                                      &                                  & $\epsilon$ = 8                                  & 47.0                               & 18.0                               & 14.0                                & 6.0                                 & \multicolumn{1}{c|}{4.0}             & 89.0                               & 54.0                               & 46.0                                & 33.0                                & \multicolumn{1}{c|}{26.0}            & 100.0                              & 100.0                              & 100.0                               & 100.0                               & 97.0                                \\
                                                      &                                  & $\epsilon$ = 16                                 & 73.0                               & 35.0                               & 30.0                                & 24.0                                & \multicolumn{1}{c|}{15.0}            & 96.0                               & 77.0                               & 66.0                                & 58.0                                & \multicolumn{1}{c|}{46.0}            & 100.0                              & 100.0                              & 100.0                               & 100.0                               & 100.0                               \\
                                                      &                                  & $\epsilon$ = 32                                 & 88.0                               & 51.0                               & 43.0                                & 31.0                                & \multicolumn{1}{c|}{26.0}            & 99.0                               & 84.0                               & 76.0                                & 69.0                                & \multicolumn{1}{c|}{64.0}            & 100.0                              & 100.0                              & 100.0                               & 100.0                               & 100.0                               \\
                                                      &                                  & $\epsilon$ = 64                                 & 95.0                               & 73.0                               & 61.0                                & 49.0                                & \multicolumn{1}{c|}{41.0}            & 100.0                              & 88.0                               & 86.0                                & 80.0                                & \multicolumn{1}{c|}{78.0}            & 100.0                              & 100.0                              & 100.0                               & 100.0                               & 100.0                               \\
                                                      &                                  & $\epsilon$ = 128                                & 99.0                               & 86.0                               & 81.0                                & 70.0                                & \multicolumn{1}{c|}{62.0}            & 100.0                              & 97.0                               & 93.0                                & 89.0                                & \multicolumn{1}{c|}{83.0}            & 100.0                              & 100.0                              & 100.0                               & 100.0                               & 100.0                               \\ \bottomrule
\end{tabular}
}
\end{table*}

\subsection{Evaluation results}
In this section, we present the evaluation results of our experiments and discuss the findings. In addition to previous charts, the main results of our experiments are reported in Table \ref{tab:ApprochesComparison}. To get these results, we crafted examples on one model per experiment then we evaluated them against all networks indepndently. To assess the effect of compression to the adversarial trigger, all the input faces are fed into networks uncompressed, and crafted adversarial examples are stored with JPEG compression. Another important aspect that we included in our investigation is the effect of resizing crafted examples. FaceNet is different from the other two networks in terms of input image size. While FaceNet accepts images with size $160\times160$, ResNet and SENet accept $224\times224$. This means that de-identified faces experience image resize when they are crafted in FaceNet as source network and classified in ResNet and SENet as target network and vice versa. 
Looking at Figures \ref{fig:PSR_Facenet}, \ref{fig:PSR_Resnet}, \ref{fig:PSR_Senet}, the first apparent understanding that spring to mind is that all adversarial examples crafted using a specific source model (FaceNet, ResNet, or SENet) transfer particularly well when considering identification based on the same recognition model. In addition, it is clear that the examples generated by FaceNet are more transferable compared to those crafted by ResNet and SENet. Comparing Figure \ref{fig:PSR_Facenet} with Figures \ref{fig:PSR_Resnet}, \ref{fig:PSR_Senet}, it can be seen that examples crafted by FaceNet using BIM method at $\epsilon = 32$ reported high transferability as they are highly protective when they were classified by the other two networks.
It is also obvious that, in all figures, when the perturbation budget increases (i.e. as the quality of the image is decreasing due to adding more noise), the protection success rate increases as well, but at the cost of sacrificing image quality. Although a higher amount makes produced examples more transferable. Considering these charts, what surprised us has been the outperformance of the BIM method which is an untargeted approach compared to the ILLC as a targeted method. Taking into account Figure \ref{fig:PSR_Facenet}, Top-25 charts, it can be noticed that while in BIM chart at $\epsilon = 32$, $PSR \geq 95\%$ for ResNet and SENet while the corresponding ones for ILLC are $PSR \leq 65\%$. 

Table \ref{tab:ApprochesComparison} shows the comparison of the protection success rate for de-identified faces crafted by BIM and ILLC adversarial methods with various noise budgets.
FaceNet outperforms other networks by protecting examples up to $89\%$ and $91\%$ on top-50 using BIM method with $\epsilon=32$ against ResNet and SENet respectively. For crafted examples at $\epsilon = 8$, which is a quite unnoticeable perturbation according to Figure \ref{fig:de-identified}, the largest protection rate is $58\%$ at Top-1 using BIM method against SENet, whereas the corresponding value using ILLC method is not higher than $32\%$. Comparing these three networks in terms of achieving higher protection success rate, ResNet reported the worst performance with $\textrm{PSR} = 39\%$ against itself for Top-1 at $\epsilon=4$ and with $\textrm{PSR} = 21\%$ against FaceNet for Top-5 at $\epsilon=32$ where the perturbation is pretty perceptible. 

These results show that using BIM and ILLC adversarial methods to preserve privacy for face images can only be achievable with $\epsilon > 32$ at the cost of degrading the quality of the image. It also indicates that the transferability, as the Protection Success Rate of the crafted examples is highly affected by resizing the examples and the difference of training parameters between source and target networks. Finally, these results point out that untargeted methods need further attention as in our experiments BIM performed better than the ILLC.

\section{Conclusion}
This paper has explored the effectiveness of adversarial methods to de-identify face biometrics: hindering automatic face recognition while preserving the visual appearance of face images. The experimental results indicate that BIM (an untargeted de-identification method) works better than ILLC (a targeted method) in terms of transferability of the crafted examples. It is likely that untargeted method are more protective than targeted ones. Yet, further studies are needed to prove this hypothesis. Besides, using these two methods, it's not possible to get a high de-identification rate with completely imperceptible perturbation. That's why most of the current literature suggests keeping the balance between the suppressing identification rate and the image quality. To this end, in our future study, we will focus on the effectiveness and transferability of less destructive adversarial methods to preserve the quality of the image including one-pixel attack, Jacobian-based Saliency Map Attack (JSMA), and deepfool \cite{moosavi2016deepfool}. We will also check the robustness of generated examples against an already trained model with adversarial examples or procedures \cite{REF10}. Finally, we will compare the crafted de-identified face images with commercial face recognition systems. 


\section{Acknowledgments}
This work has been supported by projects: PRIMA (ITN-2019-860315), TRESPASS-ETN (ITN-2019-860813), and BBforTAI (PID2021-127641OB-I00 MICINN/FEDER).

\bibliography{aaai23}



\end{document}